\documentclass[graybox]{svmult}

\usepackage{type1cm}        
\usepackage{makeidx}         
\usepackage{graphicx}        
\usepackage{multicol}        
\usepackage[bottom]{footmisc}

\usepackage{empheq}

\usepackage{newtxtext}       %
\usepackage[varvw]{newtxmath}       

\usepackage{float}
\usepackage{caption}       
\usepackage{subcaption}    

\newcommand{\Rho}{\mathrm{P}}

\DeclareMathOperator*{\argmin}{arg\,min}

\makeindex             

\begin{document}

\title*{Neural Galerkin Normalizing Flow for Transition Probability Density Functions of Diffusion Models}
\titlerunning{Neural Galerkin Normalizing Flow for Transition Probability Density Functions}
\author{Riccardo Saporiti\orcidID{0009-0005-1813-2022} and\\ Fabio Nobile\orcidID{0000-0002-8130-0114}}
\institute{R. Saporiti \at CSQI, \'Ecole Polytechnique F\'ed\'erale de Lausanne, Switzerland, \email{riccardo.saporiti@epfl.ch}
\and F. Nobile \at CSQI, \'Ecole Polytechnique F\'ed\'erale de Lausanne, Switzerland, \email{fabio.nobile@epfl.ch}}
%
%
\maketitle

\abstract*{We propose a new Neural Galerkin Normalizing Flow framework to approximate the transition probability density function of a diffusion process by solving the corresponding Fokker-Planck equation with an atomic initial distribution, parametrically with respect to the location of the initial mass. \\
By using Normalizing Flows, we look for the solution as a transformation of the transition probability density function of a reference stochastic process, ensuring that our approximation is structure-preserving and automatically satisfies positivity and mass conservation constraints. \\
 By extending Neural Galerkin schemes to the context of Normalizing Flows, we derive a system of ODEs for the time evolution of the Normalizing Flow's parameters. Adaptive sampling routines are used to evaluate the Fokker-Planck residual in meaningful locations, which is of vital importance to address high-dimensional PDEs.
\\
Numerical results show that this strategy captures key features of the true solution and enforces the causal relationship between the initial datum and the density function at subsequent times. 
\\
After completing an offline training phase, online evaluation becomes significantly more cost-effective than solving the PDE from scratch.
The proposed method serves as a promising surrogate model, which could be deployed in many-query problems associated with stochastic differential equations, like Bayesian inference, simulation, and diffusion bridge generation.
}

\abstract{We propose a new Neural Galerkin Normalizing Flow framework to approximate the transition probability density function of a diffusion process by solving the corresponding Fokker-Planck equation with an atomic initial distribution, parametrically with respect to the location of the initial mass. \\
By using Normalizing Flows, we look for the solution as a transformation of the transition probability density function of a reference stochastic process, ensuring that our approximation is structure-preserving and automatically satisfies positivity and mass conservation constraints. \\
 By extending Neural Galerkin schemes to the context of Normalizing Flows, we derive a system of ODEs for the time evolution of the Normalizing Flow's parameters. Adaptive sampling routines are used to evaluate the Fokker-Planck residual in meaningful locations, which is of vital importance to address high-dimensional PDEs.
\\
Numerical results show that this strategy captures key features of the true solution and enforces the causal relationship between the initial datum and the density function at subsequent times.
\\
After completing an offline training phase, online evaluation becomes significantly more cost-effective than solving the PDE from scratch.
The proposed method serves as a promising surrogate model, which could be deployed in many-query problems associated with stochastic differential equations, like Bayesian inference, simulation, and diffusion bridge generation.
}

\section{Introduction}

We consider the problem of solving the Fokker-Planck (FP) equation associated with the transition probability density function (TPDF) of a stochastic differential equation (SDE). The TPDF satisfies a time-dependent partial differential equation (PDE) with an atomic initial condition, namely a Dirac delta distribution centered at a specific point in the domain. 

The solution of the PDE remains highly concentrated in space at times close to the initial one, a feature that poses significant difficulties for traditional solvers that rely on spatial discretization and meshing techniques. Standard grid-based methods are notoriously sensitive to the problem's dimensionality and, in high dimensions, are likely to yield a numerical solution that is not guaranteed to be a legitimate density function. This limits their applicability in many practical scenarios.
In recent years, mesh-free approaches have become a very popular alternative for solving high-dimensional PDEs, as noted in \cite{karniadakis2021piml} and references therein. 
We consider a data-free scenario in which one lacks access to high-fidelity solutions and must rely solely on physics. In this context, Physics-Informed Neural Networks (PINNs) \cite{RAISSI2019686, SIRIGNANO20181339} have achieved spectacular results across a broad range of PDEs, relying on the universal approximation results enjoyed by Neural Networks.

In their more traditional implementation, PINNs use a space-time collocation approach that aims to minimize the PDE residual over the entire domain.  
Several works have already tackled FP-type PDEs using PINNs.
In \cite{su2024deep}, an LSTM-like architecture is used to obtain a parametric solution of the Kolmogorov Backward equation, which subsequently leads to the TPDF of the FP equation via automatic differentiation.
Temporal Normalizing Flows have also been used to approximate the solution of FP equations \cite{lu2022learning, Xiaodong_Feng2022-ak}. They express the density at each time as a transformation of a source distribution, typically set to a standard Gaussian. While \cite{lu2022learning} fits the parameters of this transformation by maximizing the likelihood function evaluated over snapshots of sample paths of the corresponding SDEs, \cite{Xiaodong_Feng2022-ak} trains the Temporal Normalizing Flow by minimizing a space-time PINN loss. 
The success of these ideas comes from the properties that the numerical solution satisfies. Notably, relying on the change of variable formula, the approximate solution is guaranteed to be a valid density function. 

 Despite its promising results, the PINN framework faces considerable challenges when applied to transport-dominated PDEs with atomic initial data. Notoriously, the crude application of this method struggles to enforce causality between the initial condition and the solution of the equation, an element of utmost importance for our application, where the location of the Dirac distribution drives the TPDF.
 Properly balancing different terms in the loss function, that is, boundary loss, initial loss, and residual of the PDE, poses additional difficulties due to the aforementioned Dirac distribution. Unless exactly satisfied, the initial loss would completely dominate the PDE residual term, limiting the model's capacity to learn the problem's physics. Another critical issue is the choice of the sampling measure used to estimate the residual in the space-time domain. A uniform distribution is likely to be affected by the curse of dimensionality and miss regions where the problem's density is evolving. 
 In this respect, Temporal Normalizing Flows are very appealing as they enable sampling from the current solution. However, this approach requires setting up adaptive loops in which the solution and the sampling measure are progressively improved over iterations. To initiate this process effectively, a reasonably accurate initial sampling measure has to be chosen, which is often not readily available.
The recent work \cite{zeng2026operatorlearningsolvingfokkerplanck}, developed in parallel to ours, addresses several of these challenges by suggesting using the TPDF of a suitably linearized process as a source distribution for the Normalizing Flow.

In this work, we follow a different approach, known as Neural Galerkin \cite{BRUNA2024112588}. It consists of deriving a system of ordinary differential equations (ODEs) to describe the time evolution of the parameters of the neural network that represents the solution of a PDE \cite{BRUNA2024112588}. 
By time integration of the ODE, the model learns the causal dependence on the time variable and can efficiently solve parametric PDEs characterized by a slow decay of the Kolmogorov n-width, such as advection-dominated problems and FP equations. 
We mention, in particular, the work \cite{doi:10.1137/20M1344986}, which uses the Wasserstein gradient flow structure of the FP equation to formulate a similar system of ODEs.
However, this method is not explicitly designed to solve the FP equation for the parametric TPDF, and is tested on a class of SDEs featuring a constant diffusion coefficient and drift induced by the gradient of a potential function.

Building on these ideas, we develop a general framework for training Normalizing Flow-based models according to the Dirac-Frenkel variational principle, yielding Neural Galerkin Normalizing Flow (NGNF) architectures. 
The generative model learns parametrically, with respect to the location of the Dirac delta distribution, the TPDF of the FP equation.
Similar to \cite{zeng2026operatorlearningsolvingfokkerplanck}, to address the degeneracy of the solution at the initial time, we propose to use a time-dependent source distribution in the Normalizing-Flow construction. However, we choose the source density to be the TPDF associated with discretizing the SDE using a single step of the Euler-Maruyama numerical scheme, which can be expressed in closed form.
The Normalizing Flow is then forced to be the identity at initial time. This allows one to obtain a solution that accurately captures how the initial density propagates during the first portion of the time integration interval. 

The methodology presented in this paper addresses the limitations associated with PINN-based Temporal Normalizing Flows by incorporating two adaptive steps. First, the architecture of the Normalizing Flow is tailored to the specific initial condition. 
Second, by using a Neural Galerkin framework, we can easily build a sampling measure sequentially in time using the currently available solution, thus avoiding unnecessary adaptive loops over the whole space-time domain and pre-training of an initial sampling measure. 

Our approach is grounded in an offline training phase that uses Neural Galerkin schemes to evolve the dynamics of the Normalizing Flow parameters. 
The computational costs incurred during this offline phase are amortized during online deployments, which only require a straightforward evaluation of the Normalizing Flow at any new location of the initial Dirac delta, with no additional training costs. The approximate TPDF can also be used online to estimate, via a suitable convolution integral, the solution of the FP equation for any arbitrary (smooth) initial condition.
 Another application is diffusion bridge generation: the surrogate model can be used to approximate the intractable drift term that guides the conditioned diffusion process towards the desired endpoint.  
The relevance of this work also lies in the importance of the transition probability density function in inference of State-Space models. In our companion paper \cite{saporiti2026neuralgalerkinnormalizingflows}, we extend the NGNF framework by training a novel class of smooth, bounded Normalizing Flows, designed to model the TPDF of SDEs associated with stochastic volatility models. 
The surrogate transition density, parametrized by the location of the initial condition and the SDE parameters, enables efficient, discretization-free Bayesian inference for popular pricing models.

\section{Problem Formulation}

Let $\boldsymbol{X}(t)$ be a $d$-dimensional Stochastic Process satisfying an SDE of the form
\begin{equation} \label{eq:ngnf_proc_Diffusion_process}
    \begin{cases}
        \begin{array}{ll}
            d\boldsymbol{X}(t) = \boldsymbol{b}(t,\boldsymbol{X}(t))dt + \sqrt{\Sigma(t,\boldsymbol{X}(t))}d\boldsymbol{W}(t) & \text{for $t\in(s,T]$}, \\
            \boldsymbol{X}(s)=\boldsymbol{x}_0,
        \end{array}
    \end{cases}
\end{equation}
where $T>s\geq0$ are fixed parameters, $\boldsymbol{b}:[s,T]\times\mathbb{R}^{d}\rightarrow\mathbb{R}^{d}$ is a drift function, $\Sigma:[s,T]\times\mathbb{R}^{d}\rightarrow\mathbb{R}^{d\times d}$ is a positive definite matrix valued function, $\boldsymbol{W}(t)$ is a $d$ dimensional vector of uncorrelated Brownian Motions and $\boldsymbol{x_0}\in\mathbb{R}^{d}$. 

The transition probability density function (TPDF) of $\boldsymbol{X}(t)$, conditioned on $\boldsymbol{X}(s)=\boldsymbol{x}_0$, is denoted by $\rho(\boldsymbol{x}|t,s,\boldsymbol{x_0})$
and satisfies $\mathbb{P}(\boldsymbol{X}(t)\in A|\boldsymbol{X}(s)=\boldsymbol{x_0})=\int_{A}\rho(\boldsymbol{x}|t,s,\boldsymbol{x_0})d\boldsymbol{x}$ for any Borel set $A\subset\mathbb{R}^{d}$. The TPDF satisfies the Fokker-Planck equation \cite{Stochastic_Processes_and_Applications}
\begin{subequations}\label{eq:ngnf_proc_TPDF_Fokker_Planck}
\begin{empheq}[left=\empheqlbrace]{align}
\label{eq:ngnf_proc_TPDF_Fokker_Planck_a}
\partial_{t}\rho(\boldsymbol{x}| t,s,\boldsymbol{x_0})
&= \mathcal{L}^{\star}_t\left(\rho(\cdot| t,s,\boldsymbol{x_0})\right)(\boldsymbol{x})
&& \text{for $\boldsymbol{x}\in\mathbb{R}^{d},\ t\in(s,T]$}, \\
\label{eq:ngnf_proc_TPDF_Fokker_Planck_b}
\rho(\boldsymbol{x}| s,s,\boldsymbol{x_0})
&= \delta_{x_0}(\boldsymbol{x})
&& \text{for $\boldsymbol{x}\in\mathbb{R}^{d}$}, \\
\label{eq:ngnf_proc_TPDF_Fokker_Planck_c}
\int_{\mathbb{R}^{d}} \rho(\boldsymbol{x}| t,s,\boldsymbol{x_0})\,\mathrm{d}\boldsymbol{x}
&= 1
&& \text{for $t\in(s,T]$}, \\
\label{eq:ngnf_proc_TPDF_Fokker_Planck_d}
\rho(\boldsymbol{x}| t,s,\boldsymbol{x_0})
&\ge 0
&&\text{for $ \boldsymbol{x}\in\mathbb{R}^{d},\ t\in(s,T]$},
\end{empheq}
\end{subequations}
where $\delta_{y}( \cdot )$ denotes the Dirac delta distribution centered at $\boldsymbol{y}\in\mathbb{R}^{d}$ and $\mathcal{L}^{\star}$ is the $L^{2}$ adjoint of the generator of the Markov Process $\boldsymbol{X}(t)$ and reads
\begin{equation} \label{eq:ngnf_proc_Adjoint_generator}
    \mathcal{L}^{\star}_t(f(\cdot))(\boldsymbol{x}) = \nabla_{\boldsymbol{x}}\cdot [ -\boldsymbol{b}(t,\boldsymbol{x})f({\boldsymbol{x}})+\frac{1}{2}\nabla_{\boldsymbol{x}}\cdot\left(\Sigma(t,\boldsymbol{x})f(\boldsymbol{x})\right) ].
\end{equation}
The TPDF is the Green's function of the FP equation, from which the solution $\rho(\boldsymbol{x}|t,s,p_0)$ to \eqref{eq:ngnf_proc_TPDF_Fokker_Planck} with an arbitrary initial distribution $p_{0}(\boldsymbol{x})$ at time $s$ can be computed by convolution; see \cite[Chapter 2.5]{Stochastic_Processes_and_Applications}:
\begin{equation} \label{eq:ngnf_proc_green_identity}
    \rho(\boldsymbol{x}|t,s,p_0) = \int_{\mathbb{R}^{d}} \rho(\boldsymbol{x}|t,s,\boldsymbol{x_0})p_{0}(\boldsymbol{x_0}) d\boldsymbol{x_0}.
\end{equation}

In the following sections, we present a framework for learning a generative model, parametrized by $\boldsymbol{x_0}\in\mathbb{R}^d$, that approximates the solution of \eqref{eq:ngnf_proc_TPDF_Fokker_Planck}. The structure of the model will guarantee that \eqref{eq:ngnf_proc_TPDF_Fokker_Planck_b}, \eqref{eq:ngnf_proc_TPDF_Fokker_Planck_c}, \eqref{eq:ngnf_proc_TPDF_Fokker_Planck_d} are naturally satisfied. The parameters of the neural network will be time-dependent functions, satisfying dynamics derived from \eqref{eq:ngnf_proc_TPDF_Fokker_Planck_a}.
 This approach implicitly defines an operator that maps, by approximating \eqref{eq:ngnf_proc_green_identity}, an arbitrary initial condition $p_0$ to the solution of the corresponding FP equation $\rho(\boldsymbol{x}|t,s,p_0)$. 


\section{Normalizing Flows for Fokker-Planck equations}\label{sec:ngnf_proc_structure_normalizing_flow}

In this section, we model the solution of \eqref{eq:ngnf_proc_TPDF_Fokker_Planck} with a parametric, time dependent approximation. At time $t\in[s,T]$, we represent the TPDF with the following ansatz
\begin{equation} \label{parametrization_TPDF}
\rho(\boldsymbol{x}|t,s,\boldsymbol{x_0})\simeq\Rho(\boldsymbol{x}|\boldsymbol{\theta}(t-s),t-s,\boldsymbol{x_0}),
\end{equation}
$\boldsymbol{\theta}(\tau)\in\Theta\subseteq\mathbb{R}^{M}$ being a time dependent vector of parameters characterizing  $\Rho:\mathbb{R}^d\times\Theta\times[0,T-s]\times\mathbb{R}^d\rightarrow\mathbb{R}$. By using the change of variable formula, we relate \eqref{parametrization_TPDF} to a reference, simpler distribution that is dependent on $\tau$, denoted as $p_{Z}(\cdot|\tau,\boldsymbol{x}_0)$

\begin{equation} \label{eq:ngnf_proc_FP_approximation_by_NF}
    \Rho(\boldsymbol{x}|\boldsymbol{\theta}(\tau),\tau,\boldsymbol{x_0})=p_{Z}(\boldsymbol{n}_{\theta(\tau)}(\boldsymbol{x}|\boldsymbol{x_0})|\tau,\boldsymbol{x_0})\lvert \mathrm{det}(\nabla_{\boldsymbol{x}}\boldsymbol{n}_{\theta(\tau)}(\boldsymbol{x}|\boldsymbol{x_0}))|,
\end{equation}
where $\boldsymbol{n}_{\theta}$ is a diffeomorphism that transforms $\Rho(\cdot|\boldsymbol{\theta}(\tau),\tau,\boldsymbol{x_0})$ into $p_{Z}(\boldsymbol{n}_{\theta(\tau)}(\cdot|\boldsymbol{x_0})|\tau,\boldsymbol{x_0})$.
The map $\boldsymbol{n}_{\theta}$ has to be learnable, cheap to evaluate, expressive, and analytically invertible. 
We use Normalizing Flows in the form of Real-NVP transformations \cite{dinh2017densityestimationusingreal,TANG2020143} to address these requirements.

\subsection{Conditional Affine Coupling layers}\label{sect:cond_aff_coupling_layer}
The transport map $\boldsymbol{n}_{\theta}$ is obtained by composition of an even number of layers $L\geq2$ such that 
\begin{equation} \label{NF_mapping}
    \begin{cases}
        \boldsymbol{x}^{L}=\boldsymbol{n}_{\theta(\tau)}(\boldsymbol{x}|\boldsymbol{x_0}), \\\boldsymbol{x}^{l}=\boldsymbol{n}_{\theta^l(\tau)}^{l}(\boldsymbol{x}^{l-1}|\boldsymbol{x_0}), \quad \text{for $l=1,...,L$},\\
        \boldsymbol{x^0}=\boldsymbol{x}.
    \end{cases}
\end{equation}

We denote by $\boldsymbol{\theta}(\tau)$ the concatenation of the parameters contained in each layer of the Flow, namely, $\boldsymbol{\theta}(\tau)=[\boldsymbol{\theta}^1(\tau);\boldsymbol{\theta}^2(\tau);...;\boldsymbol{\theta}^L(\tau)]\in\Theta$.
Each transformation $\boldsymbol{n}_{\theta}^{l}$ is constructed via an affine coupling layer. Specifically, for $m<d$, the coupling layer $\boldsymbol{n}^{l}_{\theta^l(\tau)}$ applies the following bijection to a $d$-dimensional input vector $\boldsymbol{x}^{l-1}\in\mathbb{R}^{d}$
\begin{equation} \label{eq:ngnf_proc_fp_transformation_layer_l}
    \begin{cases}
        \boldsymbol{x}^l_{1:m} = \boldsymbol{x}^{l-1}_{1:m} \\
        \boldsymbol{c}^l = (\boldsymbol{x}^{l-1}_{1:m},\boldsymbol{x_0})  \\
        \boldsymbol{x}^l_{m+1:d} = \boldsymbol{x}^{l-1}_{m+1:d} \odot \left(\mathbf{1}_{d-m} + \beta\tanh\left( \boldsymbol{s}_{\theta^l(\tau)}( \boldsymbol{c}^l)\right)\right) + \text{e}^{\boldsymbol{\xi}_{l,\theta^l(\tau)}}\odot \tanh\left(\boldsymbol{t}_{\theta^l(\tau)}(\boldsymbol{c}^l)\right),
    \end{cases}
\end{equation}
where $\odot$ is the Hadamard product, $\mathbf{1}_{d-m}=[1,1,...,1]^\top\in\mathbb{R}^{d-m}$, $\beta$ is a non-trainable scalar value, $\boldsymbol{\xi}_{l,\theta^l(\tau)}\in\mathbb{R}^{d-m}$ is a trainable vector of parameters. 
$\boldsymbol{s},\boldsymbol{t}:\mathbb{R}^{m+d}\rightarrow\mathbb{R}^{d-m}$ are nonlinear functions parametrized by $\boldsymbol{\theta}^l(\tau)$ and the hyperbolic tangent acts componentwise on $\boldsymbol{s},\boldsymbol{t}$. 
We use Neural Networks based upon GRU cells to realize the nonlinear functions $\boldsymbol{s},\boldsymbol{t}$. The rationale for this choice is that, as shown in \cite{SIRIGNANO20181339}, LSTM-like architectures are particularly effective at modeling the solution of PDEs with "sharp turns" in the initial condition.

With respect to the basic Real-NVP architectures, we have introduced the conditioning variable $\boldsymbol{c}^l$, which is formed by concatenating the idle component $\boldsymbol{x}^{l-1}_{1:m}$ and $\boldsymbol{x}_0$. 
By modulating $\boldsymbol{x}^l_{m+1:d}$ with nonlinear functions of $\boldsymbol{c}^{l}$, this parameterization allows us to learn how the solution depends upon the initial condition.
Since each coupling layer leaves the first portion of the input vector unchanged, we compose $L$ of them in an alternating fashion. In other words, the first layer updates the last $d-m$ components of the state vector while leaving the first $m$ components unchanged. We then flip the state vector and transform only its last $m$ components, conditioned on the first $d-m$ components, which have been transformed by the first layer. This pattern guarantees that variables not changed by the current coupling layer are transformed by the next one.

Building upon \eqref{NF_mapping}, the map $\boldsymbol{n}_{\theta}$ is parametric with respect to $\boldsymbol{x_0}$. Its parameters $\boldsymbol{\theta}(\tau)$, which will be explained in the following sections, are trained offline once, aiming to closely follow the temporal dynamics defined by the Fokker-Planck PDE \eqref{eq:ngnf_proc_TPDF_Fokker_Planck_a}. 
Once the map is computed, it can be evaluated cheaply during the online phase for any pair $(\boldsymbol{x}, \tau)\in \mathbb{R}^d \times [0, T-s]$, given an arbitrary initial condition $\delta_{x_0}$.
By applying the change of variable formula \eqref{eq:ngnf_proc_FP_approximation_by_NF}, we obtain the corresponding value of the TPDF at the same point.  


\subsection{Properties}\label{sec:ngnf_proc_subsection_properties}

The mapping defined in \eqref{NF_mapping} is the composition of $L$ bijections of the form in \eqref{eq:ngnf_proc_fp_transformation_layer_l}, making it a bijective transformation over $\mathbb{R}^{d}$.
Its interest, from the numerical perspective, comes from the availability of an explicit expression for the determinant of the Jacobian appearing in \eqref{eq:ngnf_proc_FP_approximation_by_NF}, which is just the product of the Jacobian of each individual transformation \eqref{eq:ngnf_proc_fp_transformation_layer_l}. The Jacobian of \eqref{eq:ngnf_proc_fp_transformation_layer_l} reads
\begin{equation} \label{matrix_Jacobian}
\nabla_{\boldsymbol{x^{l-1}}} \boldsymbol{n}_{\theta^l(\tau)}^l(\boldsymbol{x}^{l-1}) =
\begin{bmatrix}
\mathbb{I}_{m} \hspace{.8em} & \hspace{.8em} 0 \\
\nabla_{\boldsymbol{x^{l-1}_{1:m}}} (\boldsymbol{n}_{\theta^l(\tau)}^l(\boldsymbol{x}^{l-1}))_{m+1:d} 
\hspace{.8em} & \hspace{.8em}
\nabla_{\boldsymbol{x^{l-1}_{m+1:d}}} (\boldsymbol{n}_{\theta^l(\tau)}^l(\boldsymbol{x}^{l-1}))_{m+1:d}
\end{bmatrix},
\end{equation}
where $\mathbb{I}_{m}$ is the $m$-dimensional identity matrix and $\nabla_{\boldsymbol{x^{l-1}_{m+1:d}}} (\boldsymbol{n}_{\theta^l(\tau)}^l(\boldsymbol{x}^{l-1}))_{m+1:d}\in\mathbb{R}^{(d-m)\times (d-m)}$ is a diagonal matrix with diagonal elements
\begin{equation*}
    \frac{\partial (\boldsymbol{n}_{\theta^l(\tau)}^l(\boldsymbol{x}^{l-1}))_i}{\partial x^{l-1}_i} = 1 + \beta \left(\tanh\left((\boldsymbol{s}_{\theta^l(\tau)}(\boldsymbol{x}_{1:m}^{l-1},\boldsymbol{x_0}))_{i-m}\right)\right) \quad \text{for $i=m+1,...,d$}.
\end{equation*}
 
Thanks to its lower triangular structure, the determinant of \eqref{matrix_Jacobian} is simply the product of its diagonal elements. 
The determinant of interest is therefore 
\begin{equation} \label{full_determinant_NF}
\begin{split}
    \mathrm{det}(\nabla_{\boldsymbol{x}}\boldsymbol{n}_{\theta(\tau)}(\boldsymbol{x}|\boldsymbol{x_0})) =\prod_{l=1}^{L}\prod_{i=m+1}^{d} \Big\{ 1 + \beta 
\tanh\left((\boldsymbol{s}_{\theta^l(\tau)}(\boldsymbol{x}_{1:m}^{l-1},\boldsymbol{x_0}))_{i-m}\right) \Big\}.
\end{split}
\end{equation}

Furthermore, the explicit inverse of \eqref{NF_mapping} is also available; indeed, each transformation \eqref{eq:ngnf_proc_fp_transformation_layer_l} is analytically invertible (cf. \cite[Section 3.3]{dinh2017densityestimationusingreal}).
This property allows for sampling from the target random variable by
first sampling from the source distribution $\boldsymbol{z} \sim\nobreak p_Z(\cdot|\tau,\boldsymbol{x_0})$ and then applying the inverse conditional Flow $\boldsymbol{x}\sim \boldsymbol{n}_{\theta(\tau)}^{-1}(\boldsymbol{z}|\boldsymbol{x_0})$. 
 

\section{Neural Galerkin scheme for Normalizing Flows}

In this section, we derive a system of ODEs for the time evolution of the parameters of the Normalizing Flow. 
At each time instant $\tau\in[0,T-s]$, the set of parameters defines a transport map $n_{\theta(\tau)}$ that captures how the solution depends on the location of the Dirac delta distribution and on the time variable.

We start by plugging the ansatz \eqref{parametrization_TPDF} in \eqref{eq:ngnf_proc_TPDF_Fokker_Planck} and, assuming regularity of $\boldsymbol{\theta}(\tau)$, we obtain the residual function $r_{t,s}:\Theta\times\dot{\Theta}\times\mathbb{R}^d\times\mathbb{R}^d\rightarrow\mathbb{R}$ defined as
\begin{eqnarray}
r_{t,s}(\boldsymbol{\theta},\boldsymbol{\eta},\boldsymbol{x},\boldsymbol{x_0})
& = & \nabla_{\boldsymbol{\theta}}\Rho(\boldsymbol{x}|\boldsymbol{\theta}(t-s),t-s,\boldsymbol{x_0})
\cdot \boldsymbol{\eta} - \mathcal{L}^{\star}_t\left(\Rho(\cdot|\boldsymbol{\theta}(t-s),t-s,\boldsymbol{x_0})\right)(\boldsymbol{x}) \nonumber \\
& & + \left|\det\left(\nabla_{\boldsymbol{x}}\boldsymbol{n}_{\theta(t-s)}(\boldsymbol{x}|\boldsymbol{x_0})\right)\right|
\partial_{t}p_{Z}\left(\boldsymbol{n}_{\theta(t-s)}(\boldsymbol{x}|\boldsymbol{x_0})|t-s,\boldsymbol{x_0}\right),
\label{eq:ngnf_proc_residual_Neural_Galerkin}
\end{eqnarray}
$\dot{\Theta}$ being the set of time derivatives of $\boldsymbol{\theta}$.
Since we aim to minimize \eqref{eq:ngnf_proc_residual_Neural_Galerkin} w.r.t. $\boldsymbol{\eta}$ for all initial conditions $\boldsymbol{x_0}\in\mathbb{R}^d$ and all points in the space domain $\boldsymbol{x}\in\mathbb{R}^d$, we evolve the parameters of the Normalizing Flow by solving the following optimization problem 
\begin{equation} \label{eq:ngnf_proc_optimization_Neural_Galerkin}
    \boldsymbol{\dot{\theta}}(\tau) \in \argmin_{\boldsymbol{\eta}\in\dot{\Theta}} J_{\tau}(\boldsymbol{\theta},\boldsymbol{\eta}), 
\end{equation}

\begin{equation} \label{eq:ngnf_proc_J_cost_functional_Neural_Galerkin}
    J_{\tau}(\boldsymbol{\theta},\boldsymbol{\eta})=\int_{\mathbb{R}^d}\int_{\mathbb{R}^d} \lvert r_{s+\tau,s}(\boldsymbol{\theta},\boldsymbol{\eta},\boldsymbol{x},\boldsymbol{x_0}) \rvert ^{2} d\nu_{{\theta}(\tau)}(\boldsymbol{x}|\boldsymbol{x_0}) d\mu(\boldsymbol{x_0}),
\end{equation}
where $\nu_{{\theta}(\tau)}$ and $\mu$ are positive measures with support over $\mathbb{R}^d$.
While prior knowledge of the model's deployment field guides the choice of $\mu$, picking $\nu_{{\theta}(\tau)}$ is arbitrary and requires particular care. Since the solution has local features, naively evaluating the residual \eqref{eq:ngnf_proc_residual_Neural_Galerkin} over uniformly sampled points in $\mathbb{R}^d$ would likely yield poor performance. The situation is worsened in a high dimensional setting, where the curse of dimensionality adds an additional burden, and the variance of a sample estimator of \eqref{eq:ngnf_proc_J_cost_functional_Neural_Galerkin} may grow unboundedly.
To evaluate the residual \eqref{eq:ngnf_proc_residual_Neural_Galerkin} efficiently, 
it is necessary to use a measure that tracks the evolution of the true solution.
While sampling from the SDE dynamics \eqref{eq:ngnf_proc_Diffusion_process} could be a viable option, this turned out to be unsatisfactory because it overlooks the variations of the numerical solution over time.
Better performance is obtained by setting $\nu_{{\theta}(\tau)}(\cdot|\boldsymbol{x_0})=\mathrm{P}(\cdot|\boldsymbol{\theta}(\tau),\tau,\boldsymbol{x_0})$ as the time-dependent density of the Normalizing Flow conditioned on $\boldsymbol{x_0}$. Samples from $\nu_{\theta}$ can be drawn by following the procedure described at the end of Sect.~\ref{sec:ngnf_proc_subsection_properties}, which relies on the explicit inverse of $\boldsymbol{n}_{\theta}$.


\subsection{Monte Carlo estimator of the parameter update equation}

We derive a Monte Carlo estimator of \eqref{eq:ngnf_proc_J_cost_functional_Neural_Galerkin} by sampling $N$ points from $\mu$, specifically $\{\boldsymbol{x_{0}}_{,i}\}_{i=1}^{N}$, and, conditioning on each of them, one point from $\nu_{{\theta}}(\cdot|\boldsymbol{x_{0}}_{,i})$, yielding $\{\boldsymbol{x}_i,\boldsymbol{x_{0}}_{,i}\}_{i=1}^{N}$.

The minimization problem described in \eqref{eq:ngnf_proc_optimization_Neural_Galerkin} is approximated as follows
\begin{equation} \label{eq:ngnf_proc_MC_optimality_condition}
    \boldsymbol{\dot{\theta}}(\tau)\in\argmin_{\boldsymbol{\eta}\in\dot{\Theta}}\hat{J}_{\tau}(\boldsymbol{\theta},\boldsymbol{\eta}) = \frac{1}{N} \|\mathrm{J}_{\tau}\boldsymbol{\eta}-\boldsymbol{\mathrm{f}}_{\tau}\|_{2}^{2},
\end{equation}
where $\mathrm{J}_{\tau}\in\mathbb{R}^{N\times M}$ is the discrete neural network Jacobian 
\begin{equation} \label{eq:ngnf_proc_NN_Jacobian}
    (\mathrm{J}_{\tau})_{ik} = \frac{\partial \Rho(\boldsymbol{x}_{i}|\boldsymbol{\theta}(\tau),\tau,\boldsymbol{x_{0}}_{,i})}{\partial \theta_k}
\end{equation}
 and $\boldsymbol{\mathrm{f}}_{\tau}\in\mathbb{R}^{N}$ is such that 
 \begin{eqnarray}
    (\mathrm{f_{\tau}})_{i}
    & = & \mathcal{L}^{\star}_{s+\tau}\left(\Rho(\cdot|\boldsymbol{\theta}(\tau),\tau,\boldsymbol{x_{0}}_{,i})\right)(\boldsymbol{x}_i) \nonumber \\
    & & + \left|\det\left(\nabla_{\boldsymbol{x}}\boldsymbol{n}_{\theta(\tau)}(\boldsymbol{x}_i|\boldsymbol{x_{0}}_{,i})\right)\right|
    \partial_{t}p_{Z}\left(\boldsymbol{n}_{\theta(\tau)}(\boldsymbol{x}_i|\boldsymbol{x_{0}}_{,i})|\tau,\boldsymbol{x_{0}}_{,i}\right).
\label{eq:ngnf_proc_singe_entree_bold_f}
\end{eqnarray}




\subsection{Imposition of the initial condition and choice of the source distribution} \label{section_initial_condition}

The minimization problem \eqref{eq:ngnf_proc_optimization_Neural_Galerkin} is complemented by an initial condition on $\boldsymbol{\theta}(0)$.
By carefully selecting the source distribution $p_{Z}$, we manage to impose \eqref{eq:ngnf_proc_TPDF_Fokker_Planck_b} exactly. 
This process is realized in two steps
\begin{enumerate}
    \item We initialize $\boldsymbol{\theta}(0)$ so that the
    transport map $\boldsymbol{n}_{\theta(0)}$ acts as the identity 
    \begin{equation} \label{eq:ngnf_proc_FP_approximation_by_NF_id}
        \Rho(\boldsymbol{x}|\boldsymbol{\theta}(0),0,\boldsymbol{x_0})=p_{Z}(\boldsymbol{x}|0,\boldsymbol{x_0}).
    \end{equation}
    \item We choose a source distribution satisfying $p_{Z}(\cdot|0,\boldsymbol{x_0})=\delta_{x_0}(\cdot)$.
\end{enumerate}

To ensure that $\boldsymbol{n}_{\theta(\tau)}$ is an identity for $\tau=0$, we initialize the parameters $\boldsymbol{\theta}^l(\tau)$ sparsely so that scaling and translation functions $\boldsymbol{s}_{\theta^l(0)}$ and $\boldsymbol{t}_{\theta^l(0)}$ both equal to $\boldsymbol{0}$ for $l=1,...,L$.
Under this initialization, each affine coupling layer performs the following transformation
\begin{equation} \label{transformation_identity_layer_l}
    \boldsymbol{x}^{l} = \boldsymbol{n}^{l}_{\theta^l(0)}(\boldsymbol{x}^{l-1}), \quad
    \begin{cases}
        \boldsymbol{x}^l_{1:m} = \boldsymbol{x}^{l-1}_{1:m} \\
        \boldsymbol{x}^l_{m+1:d} = \boldsymbol{x}^{l-1}_{m+1:d} \odot \mathbf{1}_{d-m},
    \end{cases}
\end{equation}
which preserves the volume.

Next, we define $p_{Z}(\cdot|t-s,\boldsymbol{x_0})$ as the TPDF induced by a single step of the Euler-Maruyama (EM) discretization of \eqref{eq:ngnf_proc_Diffusion_process}. Given a starting point $\boldsymbol{x_0}$, this numerical integration scheme approximates the solution of \eqref{eq:ngnf_proc_Diffusion_process} as 
\begin{equation} \label{eq:ngnf_proc_euler_discretization}
    \boldsymbol{X}^{EM}(t)|(\boldsymbol{X}^{EM}(s)=\boldsymbol{x_0}) = \boldsymbol{x_0} + \boldsymbol{b}(s,\boldsymbol{x_0})(t-s) + \sqrt{\Sigma(s,\boldsymbol{x_0})}\boldsymbol{W}(t-s),
\end{equation}
whence the distribution of $\boldsymbol{X}^{EM}(t)|(\boldsymbol{X}^{EM}(s)=\boldsymbol{x_0})$ is a scaled and shifted $d$ dimensional Gaussian distribution 
\begin{equation}\label{eq:ngnf_proc_source_density}
    \boldsymbol{X}^{EM}(t)|(\boldsymbol{X}^{EM}(s)=\boldsymbol{x_0})\sim\mathcal{N}_{d}\left(\boldsymbol{x_0} + \boldsymbol{b}(s,\boldsymbol{x_0})(t-s),\Sigma(s,\boldsymbol{x_0})(t-s)\right).
\end{equation}
For elliptic SDEs, the Gaussian distribution defined in \eqref{eq:ngnf_proc_source_density} has a nondegenerate covariance matrix. Its convergence to the true solution of \eqref{eq:ngnf_proc_TPDF_Fokker_Planck}, as $t\rightarrow s$, is theoretically established, see, e.g., \cite{kloeden1992numerical}.


\section{Numerical Experiments}
We validate our methodology through two experiments: the two-dimensional Beneš SDE as a multimodal benchmark and a sheared anisotropic Brownian motion as a high-dimensional test case. For numerical stability, motivated by \cite{doi:10.1137/24M1638768}, we formulate the residual \eqref{eq:ngnf_proc_J_cost_functional_Neural_Galerkin} in terms of the log-density $q=\log(\Rho)$, instead of directly in terms of $\Rho$.
\subsection{Beneš SDE}\label{sect:ngnf_proc_benes}
We consider the two dimensional anisotropic Beneš SDE, which, for an arbitrary rotation matrix $\mathrm{R}_{2}\in\mathbb{R}^{2\times 2}$, follows \eqref{eq:ngnf_proc_Diffusion_process} with $\boldsymbol{b}(t,\boldsymbol{x})=\mathrm{R}_{2}\tanh(\mathrm{R}_{2}^\top\boldsymbol{x})$ and $\sqrt{\Sigma(t,\boldsymbol{x})}=\mathrm{R}_{2}$.
For $\mathrm{R}_{2}=\mathbb{I}_{2}$, the solution of \eqref{eq:ngnf_proc_TPDF_Fokker_Planck} can be derived explicitly
\begin{equation} \label{eq:ngnf_proc_explicit_benes_identity}
    \rho_{\mathbb{I}}(\boldsymbol{x}|t,s,\boldsymbol{x_0})=\frac{\exp\{-(t-s)\}}{2\pi(t-s)}\frac{\cosh(x_1)\cosh(x_2)}{\cosh(x_{0,1})\cosh(x_{0,2})}\exp\left\{-\frac{\lvert\boldsymbol{x}-\boldsymbol{x}_0\rvert^2}{2(t-s)}\right\}.
\end{equation}
The solution associated with a generic $\mathrm{R}_{2}$ follows from the change of variable formula 
\begin{equation} \label{eq:ngnf_proc_explicit_benes_rotated}
    \rho_{\mathrm{R}_2}(\boldsymbol{x}|t,s,\boldsymbol{x_0})=\rho_{\mathbb{I}}(\mathrm{R}_{2}^{\top}\boldsymbol{x}|t,s,\mathrm{R}_{2}^{\top}\boldsymbol{x_0}).
\end{equation}
In the numerical experiment, we consider a rotation of $\pi/3$ by setting
\begin{equation*}
    {\mathrm{R}}_2=\begin{bmatrix}
1/2 & -\sqrt{3}/2 \\
\sqrt{3}/2 & 1/2
\end{bmatrix}.
\end{equation*}
Following the discussion of Sect.~\ref{section_initial_condition}, 
we pick the source distribution as
\begin{equation} \label{eq:ngnf_proc_source_benes}
    p_{Z}(\cdot|t-s,\boldsymbol{x_0})\sim\mathcal{N}_2(\boldsymbol{x_0} + {\mathrm{R}}_2\tanh(\mathrm{R}_{2}^\top\boldsymbol{x}_0)(t-s),(t-s)\mathbb{I}_2).
\end{equation}
The true solution is anisotropic, multimodal, and non-Gaussian. This test case aims to show that our generative model is capable of transforming the unimodal source distribution \eqref{eq:ngnf_proc_source_benes} into the correct density function.

We solve \eqref{eq:ngnf_proc_TPDF_Fokker_Planck} over the time horizon $(s,T]=(0,3]$. To capture the parametric dependence on $\boldsymbol{x_0}$, we sample the location of the initial condition from a normal distribution with standard deviation 0.75. Specifically, we choose $\mu=\mathcal{N}_{2}(\boldsymbol{0},0.75^2\mathbb{I}_2)$ in \eqref{eq:ngnf_proc_J_cost_functional_Neural_Galerkin}. 

The problem is posed in a 2+1+2 space-time-parameter domain, which, combined with the atomic initial condition, renders it computationally challenging.
In the transformation introduced in Sect.~\ref{sec:ngnf_proc_structure_normalizing_flow}, 
we set the number of coupling layers to 10 and the hidden size of the GRU cells, used to obtain the shift and the scaling of each layer of the Normalizing Flow, to 4. By posing $m=1$, we partition the components of the two dimensional input vector $\boldsymbol{x}$ into two parts.
We integrate the dynamics of ${\boldsymbol{\dot{\theta}}}(\tau)$, estimated by \eqref{eq:ngnf_proc_MC_optimality_condition}, with an explicit Runge-Kutta method of order 3(2). 

\begin{figure}[h!]
\centering
\captionsetup[subfigure]{labelformat=empty} 
\begin{subfigure}{0.3\textwidth}
  \includegraphics[width=\linewidth]{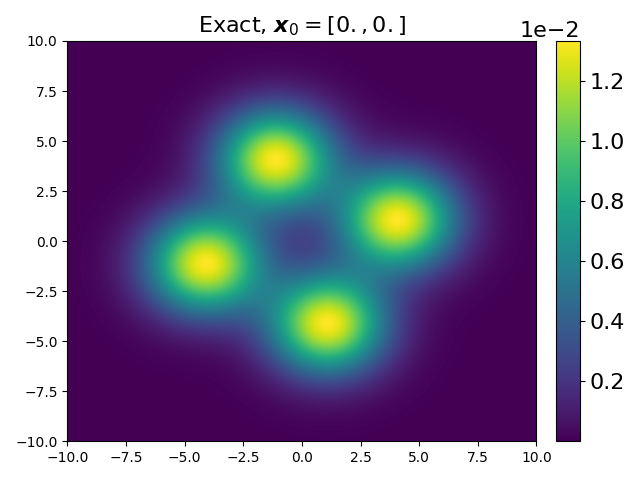}
  \caption{}
\end{subfigure}
\begin{subfigure}{0.3\textwidth}
  \includegraphics[width=\linewidth]{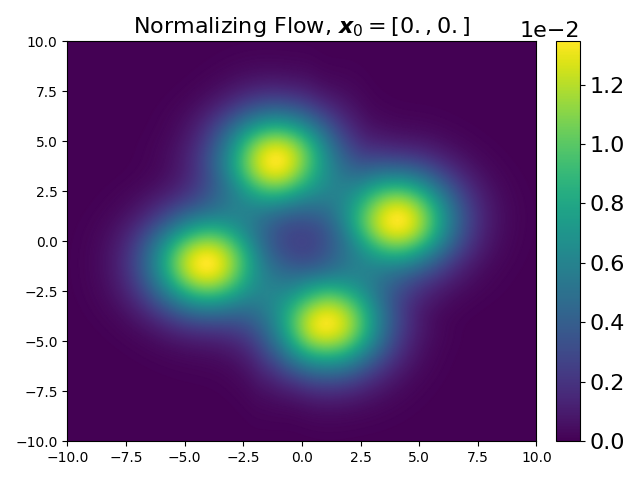}
  \caption{}
\end{subfigure}
\begin{subfigure}{0.3\textwidth}
  \includegraphics[width=\linewidth]{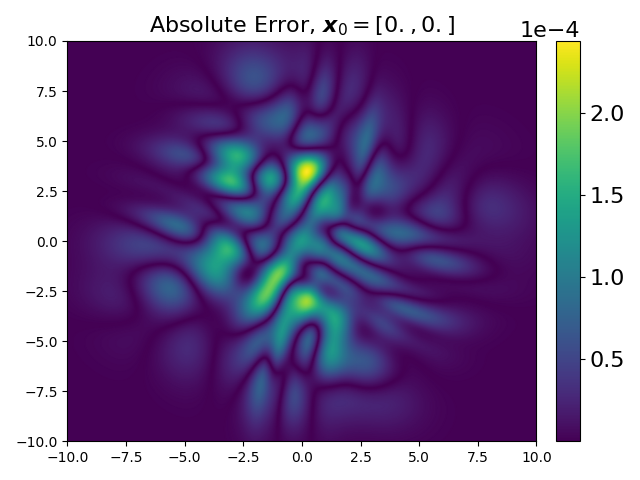}
  \caption{}
\end{subfigure}
\begin{subfigure}{0.3\textwidth}
  \includegraphics[width=\linewidth]{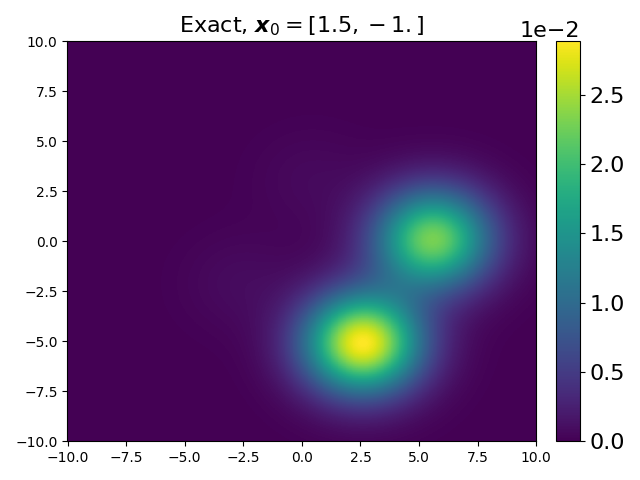}
  \caption{}
\end{subfigure}
\begin{subfigure}{0.3\textwidth}
  \includegraphics[width=\linewidth]{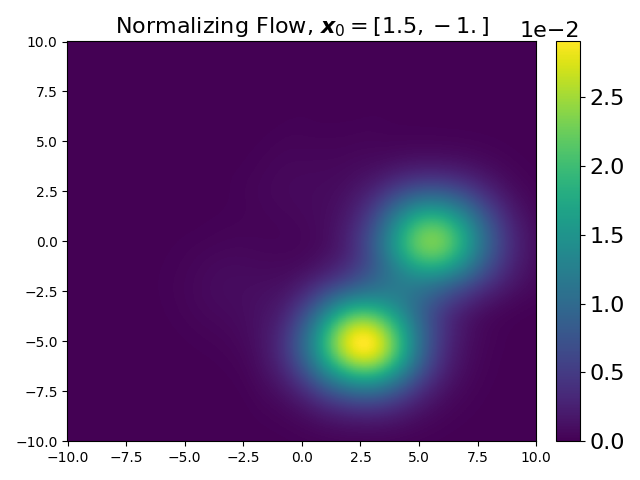}
  \caption{}
\end{subfigure}
\begin{subfigure}{0.3\textwidth}
  \includegraphics[width=\linewidth]{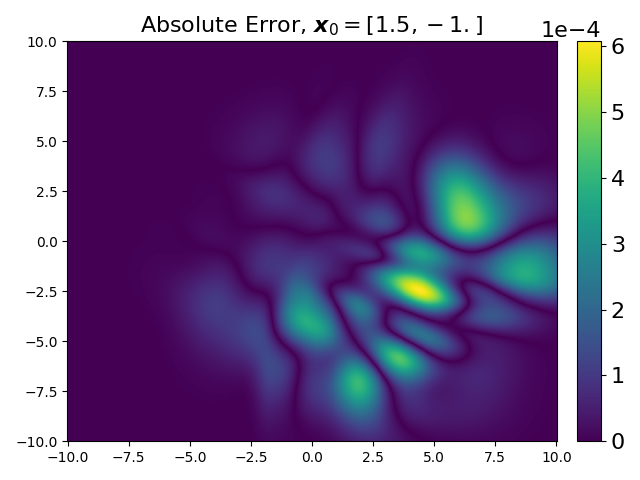}
  \caption{}
\end{subfigure}
\caption{Joint TPDF of the Beneš model associated with $\boldsymbol{x_0}=[0,0]$ (first line) and with $\boldsymbol{x_0}=[1.5,-1]$ (second line). The solution is computed at time $t=3$.}
\label{fig:ngnf_proc_NF_vs_exact_joint_benes}
\end{figure}

In Fig.~\ref{fig:ngnf_proc_NF_vs_exact_joint_benes}, we show, for different initial conditions, the true and the approximated joint density function of the Beneš model at $T=3$. 
The Normalizing Flow learns how the Dirac delta mass evolves over time and how the final solution depends on its original location. 

\begin{figure}[h!]
\centering
\captionsetup[subfigure]{labelformat=empty} 
\begin{subfigure}{0.2\textwidth}
  \includegraphics[width=\linewidth]{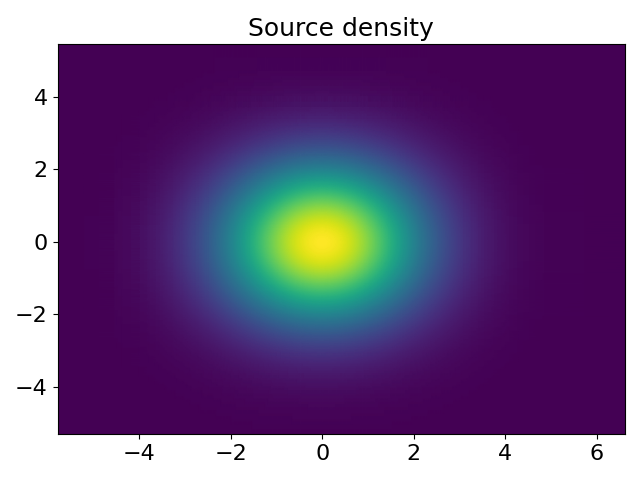}
  \caption{}
\end{subfigure}
\begin{subfigure}{0.2\textwidth}
  \includegraphics[width=\linewidth]{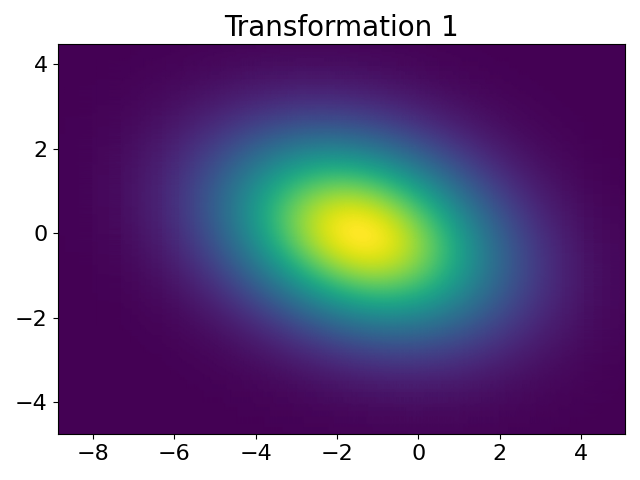}
  \caption{}
\end{subfigure}
\begin{subfigure}{0.2\textwidth}
  \includegraphics[width=\linewidth]{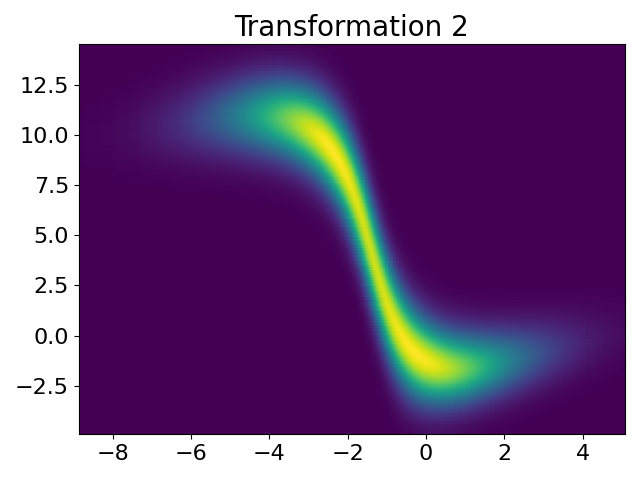}
  \caption{}
\end{subfigure}
\begin{subfigure}{0.2\textwidth}
  \includegraphics[width=\linewidth]{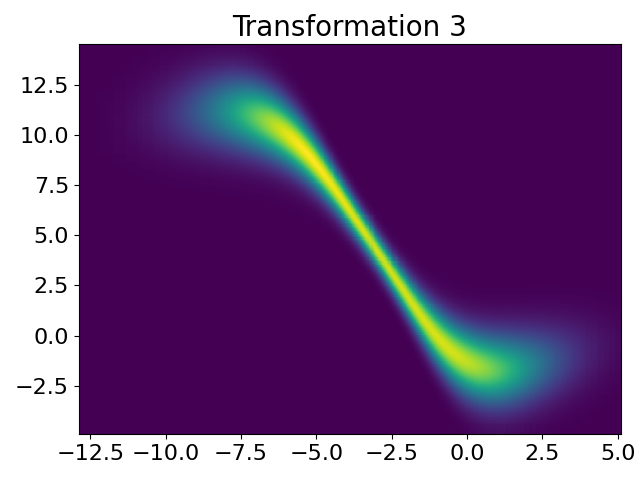}
  \caption{}
\end{subfigure}
\begin{subfigure}{0.2\textwidth}
  \includegraphics[width=\linewidth]{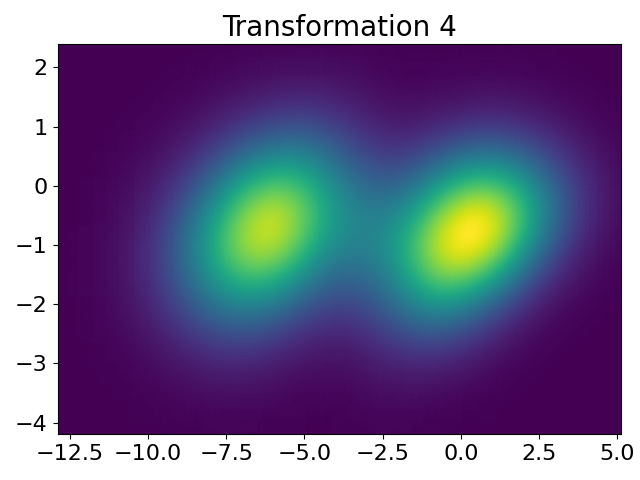}
  \caption{}
\end{subfigure}
\begin{subfigure}{0.2\textwidth}
  \includegraphics[width=\linewidth]{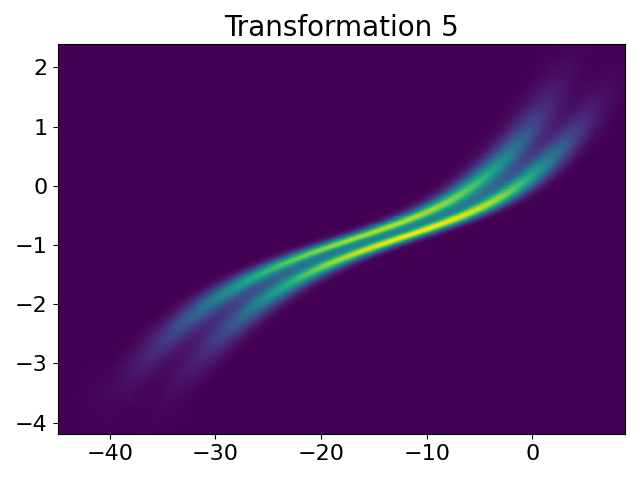}
  \caption{}
\end{subfigure}
\begin{subfigure}{0.2\textwidth}
  \includegraphics[width=\linewidth]{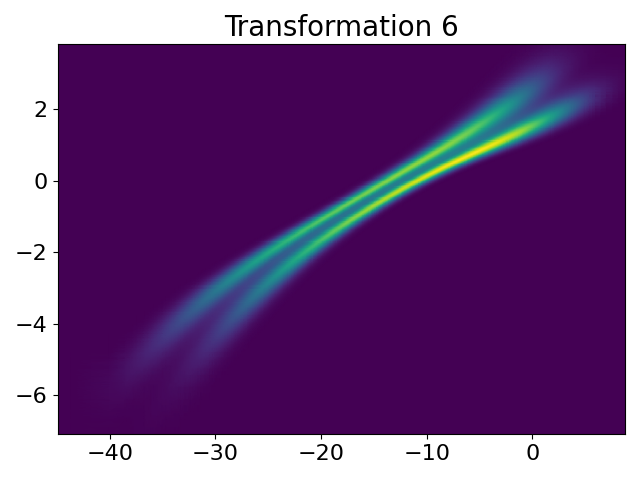}
  \caption{}
\end{subfigure}
\begin{subfigure}{0.2\textwidth}
  \includegraphics[width=\linewidth]{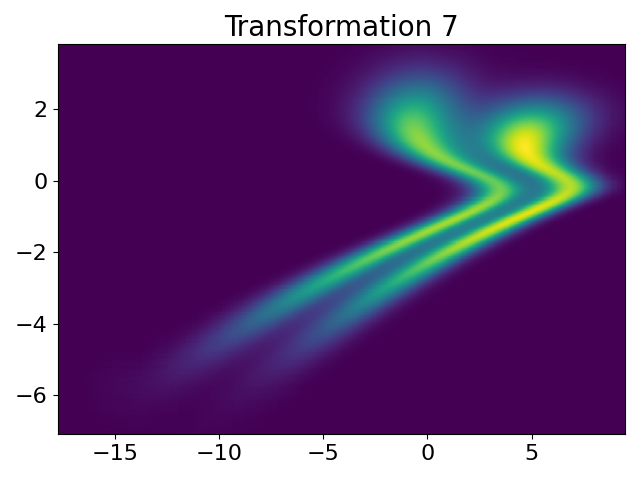}
  \caption{}
\end{subfigure}
\begin{subfigure}{0.2\textwidth}
  \includegraphics[width=\linewidth]{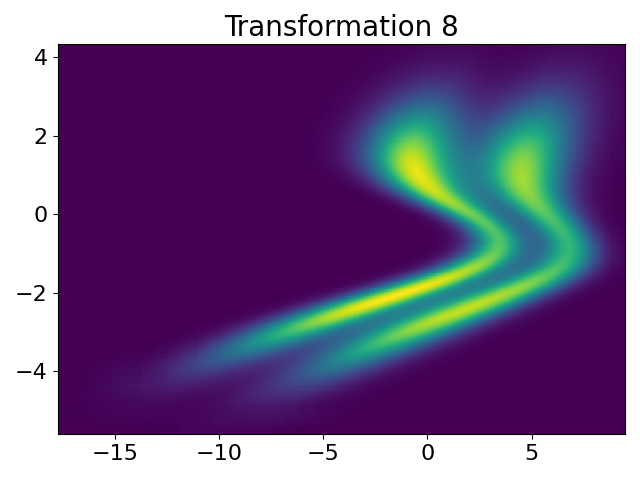}
  \caption{}
\end{subfigure}
\begin{subfigure}{0.2\textwidth}
  \includegraphics[width=\linewidth]{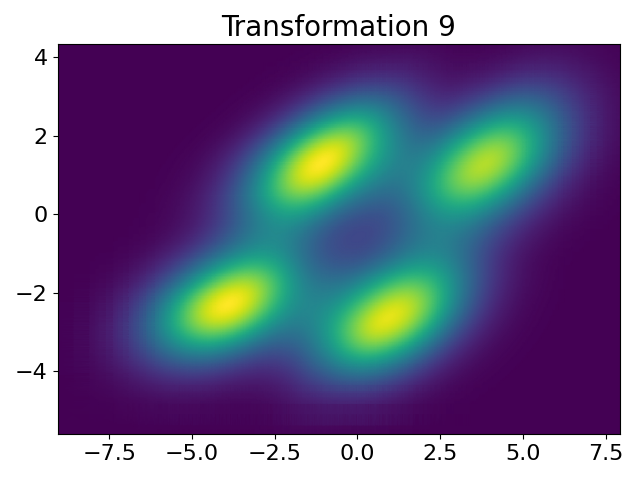}
  \caption{}
\end{subfigure}
\begin{subfigure}{0.2\textwidth}
  \includegraphics[width=\linewidth]{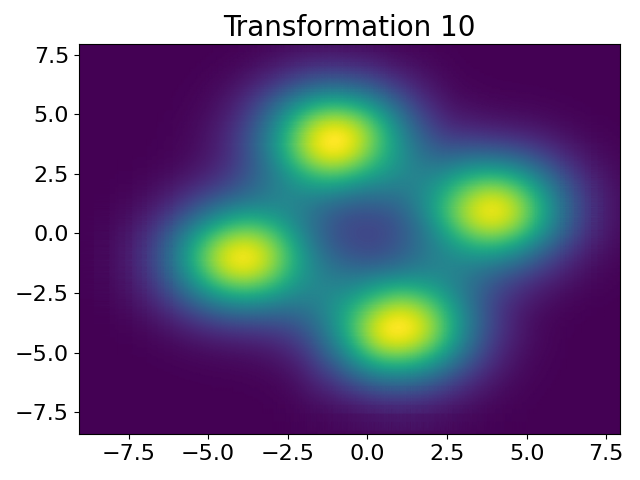}
  \caption{}
\end{subfigure}
\caption{10 layer flow molding samples from the reference Gaussian density \eqref{eq:ngnf_proc_source_benes} at $t=3$ into the density of the Beneš model at $t=3$. The initial condition is set to $\boldsymbol{x_0}=\boldsymbol{0}$.}
\label{fig:ngnf_proc_NF_transformation}
\end{figure}

To interpret the mechanism of our generative model, we show in Fig.~\ref{fig:ngnf_proc_NF_transformation} how the Flow manipulates the reference distribution for $\boldsymbol{x_0}=[0,0]$ across its 10 layers to arrive at the target density. The source distribution \eqref{eq:ngnf_proc_source_benes}, evaluated at time $T=3$, is progressively rotated and separated into the four modes of the joint density of the Beneš model. We observe that, as explained in Sect.~\ref{sect:cond_aff_coupling_layer}, the layers alternate between vertical and horizontal transformations in the corresponding latent space. 
Despite the role of some layers seeming to be redundant, this may not be the case for a different parameter $\boldsymbol{x_0}$.

\begin{figure}[t]
\centering
\begin{tabular}{@{}c c c@{}}
\includegraphics[width=0.31\textwidth]{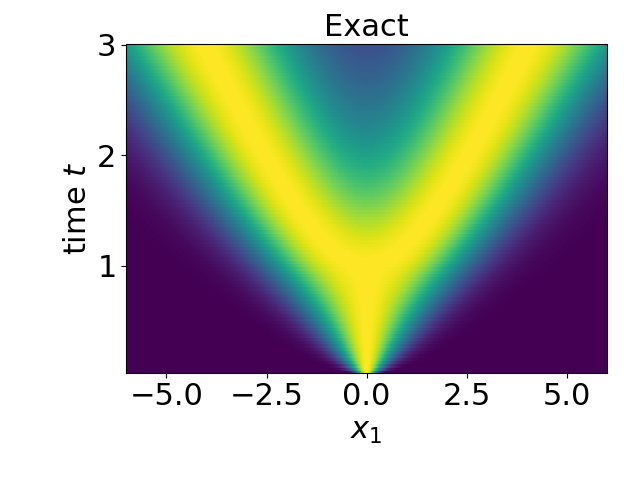} &
\includegraphics[width=0.31\textwidth]{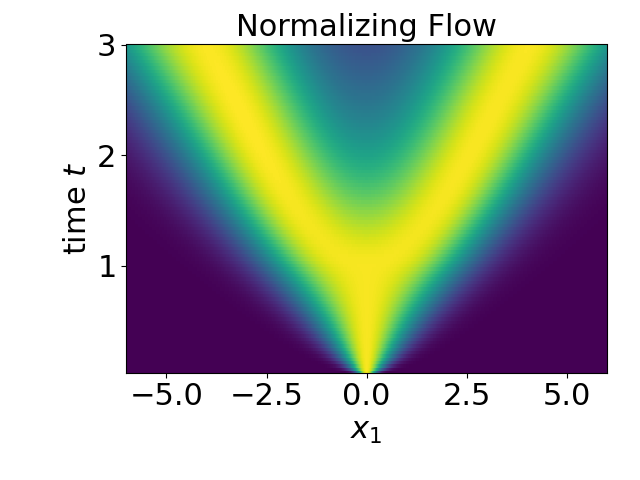} &
\includegraphics[width=0.31\textwidth]{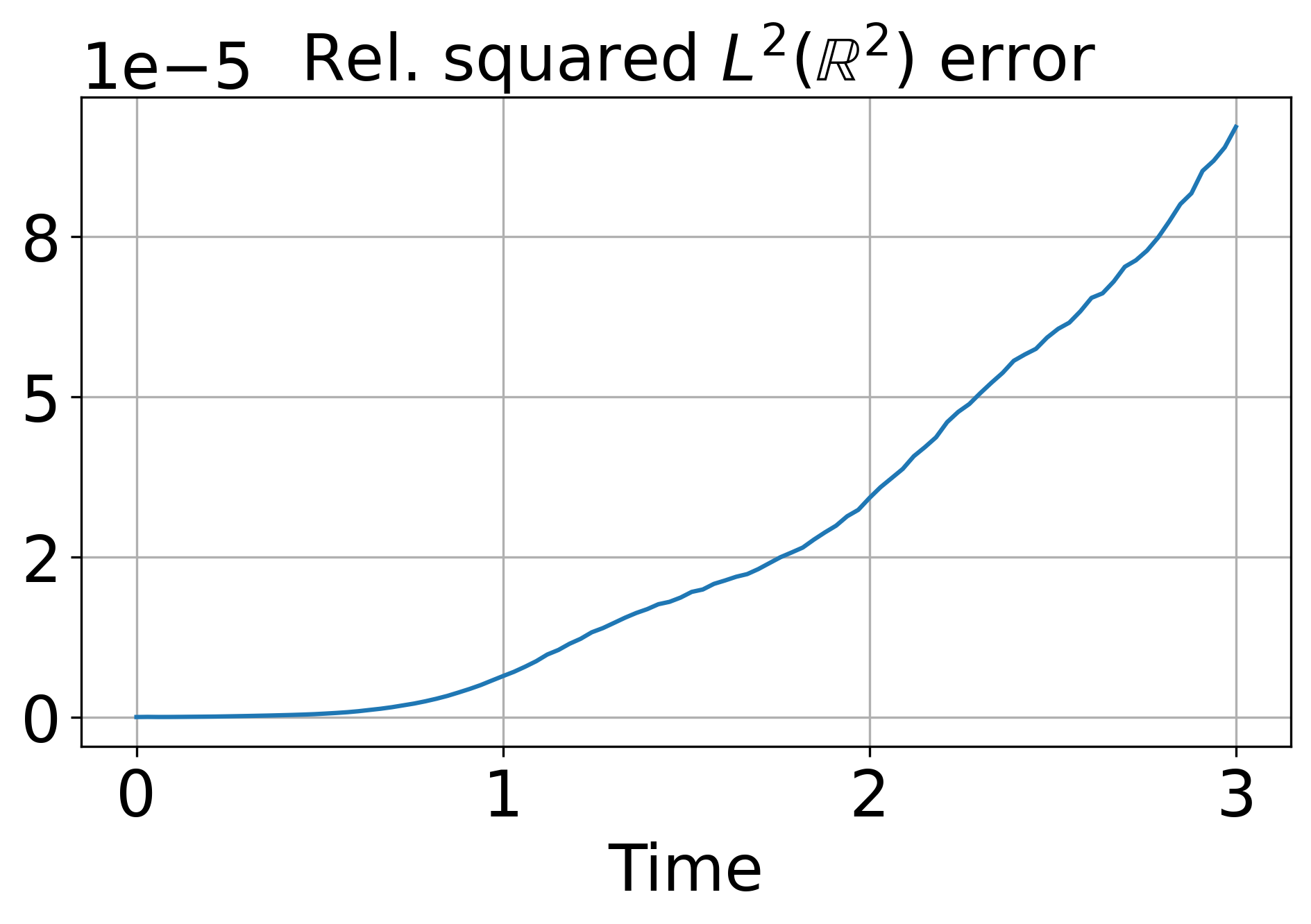} \\
\footnotesize (a) Exact conditional density & \footnotesize (b) Normalizing Flow & \footnotesize (c) 
\end{tabular}
\caption{Figure (a) and (b) show the time evolution of the exact and approximated conditional density  $\rho_{\mathrm{R}_2}(x_1|x_2=0,t,s=0,\boldsymbol{x_0}=0)$. To prevent the Dirac initial condition from dominating the plot, we rescaled $\rho_{\mathrm{R}_2}$ between 0 and 1. Figure (c) shows the time evolution of the relative (rel.) squared $L^2$ error between the true and the numerical joint density function.} 
\label{fig:ngnf_proc_NF_vs_Exact_conditional_benes}
\end{figure}

In Fig.~\ref{fig:ngnf_proc_NF_vs_Exact_conditional_benes}, we show the time evolution of the density function of the Beneš model conditioned upon $x_2=0$ and of the relative squared $L^2$ error, this time between the two joint distributions, computed over the space domain.
The relative squared $L^2$ error grows in a controlled manner, indicating that our approximation remains stable over relatively long time horizons. This stability is particularly relevant for data assimilation applications, where measurements are gathered at low acquisition frequencies, and the model must be evaluated at the time of observation. 

\begin{figure}[h!]
\centering
\captionsetup[subfigure]{labelformat=empty}
\begin{subfigure}{0.3\textwidth}
  \includegraphics[width=\linewidth]{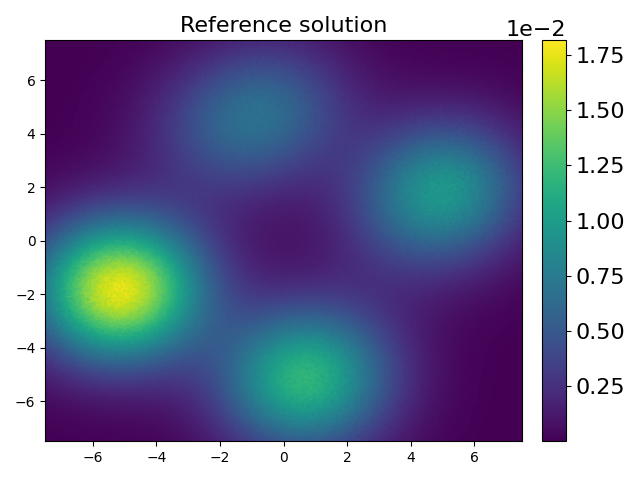}
  \caption{}
\end{subfigure}
\begin{subfigure}{0.3\textwidth}
  \includegraphics[width=\linewidth]{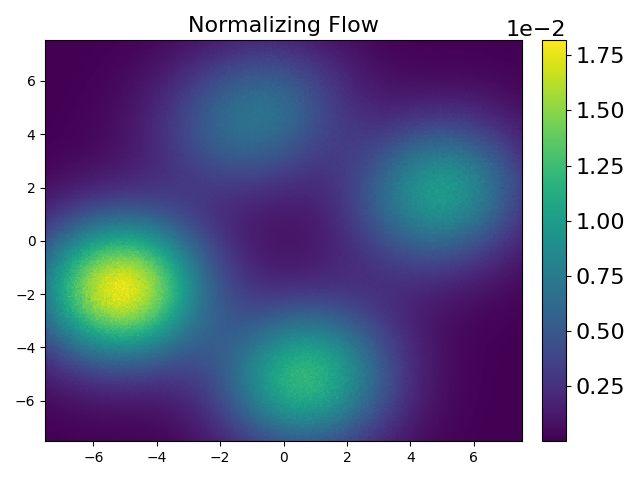}
  \caption{}
\end{subfigure}
\begin{subfigure}{0.3\textwidth}
  \includegraphics[width=\linewidth]{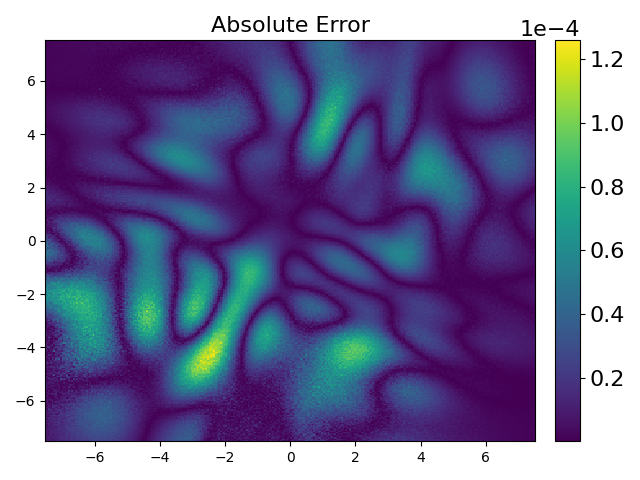}
  \caption{}
\end{subfigure}
\caption{Joint density function of the Beneš model at $t=3$ starting from the density \eqref{eq:ngnf_proc_mog_initial_cond}.
}
\label{fig:ngnf_proc_NF_vs_exact_mog_joint_benes}
\end{figure}

Finally, we evaluate the quality of our approximation by using \eqref{eq:ngnf_proc_green_identity} to compute the solution of the FP equation with the initial condition
\begin{equation} \label{eq:ngnf_proc_mog_initial_cond}
    p_0(\boldsymbol{x}) = \frac{1}{4}\phi(\boldsymbol{x};[1;1],0.1^2\mathbb{I}_2)+\frac{3}{4}\phi(\boldsymbol{x};[-0.75;-0.75],0.5^2\mathbb{I}_2),
\end{equation}
where $\phi(\cdot;\boldsymbol{\mu},\Sigma)$ is the density of a multivariate Gaussian random variable with mean $\boldsymbol{\mu}$ and covariance matrix $\Sigma$. To evaluate the distribution in each spatial point $\boldsymbol{x}$, we draw $1750$ samples from \eqref{eq:ngnf_proc_mog_initial_cond} and replace \eqref{eq:ngnf_proc_green_identity} by its Monte Carlo estimator. 
This test, which conditions our model on numerous independent and identically distributed samples, verifies the generalization capabilities of the Normalizing Flow as the location of the Dirac delta varies.
In Fig.~\ref{fig:ngnf_proc_NF_vs_exact_mog_joint_benes}, we show the joint distribution obtained by approximating the two dimensional integral \eqref{eq:ngnf_proc_green_identity}. 
Our approximation accurately captures the skewness of the solution and matches the target distribution with high accuracy.

\subsection{Sheared anisotropic Brownian motion}\label{sect:ngnf_proc_shear}

We consider a non-Gaussian transformation of the anisotropic Brownian motion introduced in \cite{doi:10.1137/24M1638768}. 
Let $B\in\mathbb{R}^{6\times 6}$ be chosen as in \cite{doi:10.1137/24M1638768}, so that $BB^\top$ is full-rank and anisotropic, and let $\boldsymbol{Y}(t)=\boldsymbol{y_0}+ B\boldsymbol{W}(t)$, where $\boldsymbol{y_0}\in\mathbb{R}^6$. 
To obtain a non-Gaussian TPDF, we transform $\boldsymbol{Y}(t)$ into $\boldsymbol{X}(t)=\boldsymbol{Y}(t)+0.75 {Y}_1^2(t)\boldsymbol{e}_2$, where 
$\boldsymbol{e}_2$ denotes the second canonical basis vector of $\mathbb{R}^6$.
We then tackle the FP PDE in a 6+1+6 space+time+parameter domain associated with $\boldsymbol{X}(t)$, 
whose SDE reads 
\begin{equation}\label{eq:ngnf_proc_SDE_high_dimension}
    d\boldsymbol{X}(t)=0.75\|\boldsymbol{e}_1^{\top}B\|^2\boldsymbol{e}_2dt+(\mathbb{I}_6+1.5X_1(t)\boldsymbol{e}_2\boldsymbol{e}_1^\top)Bd\boldsymbol{W}(t).
\end{equation}

We maintain the experimental setup as in Sect.~\ref{sect:ngnf_proc_benes}, specifically setting $m=3$ in the coupling layers. We solve the FP equation over the time horizon $(s,T]=(0,1]$.

\begin{figure}[!htbp]
\centering
\captionsetup[subfigure]{labelformat=empty}
\begin{subfigure}{0.3\textwidth}
  \includegraphics[width=\linewidth]{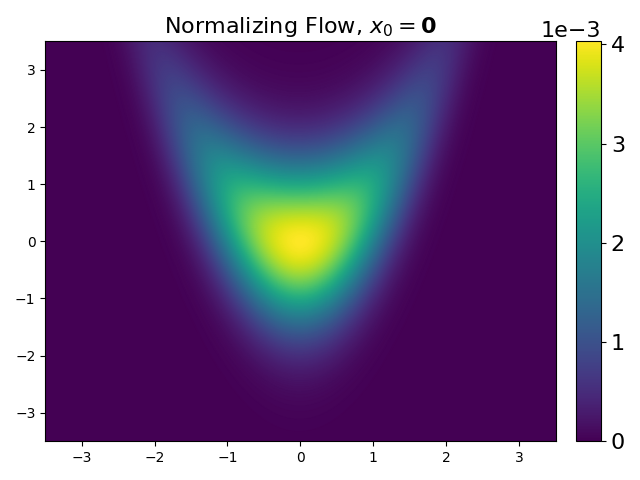}
  \caption{}
\end{subfigure}
\begin{subfigure}{0.3\textwidth}
  \includegraphics[width=\linewidth]{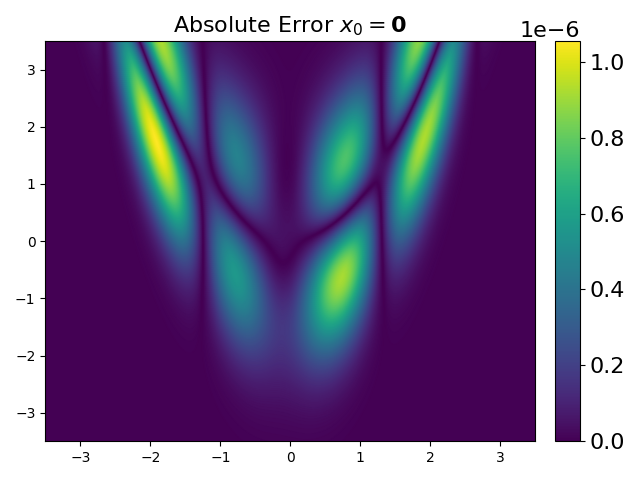}
  \caption{}
\end{subfigure}
\begin{subfigure}{0.3\textwidth}
  \includegraphics[width=\linewidth]{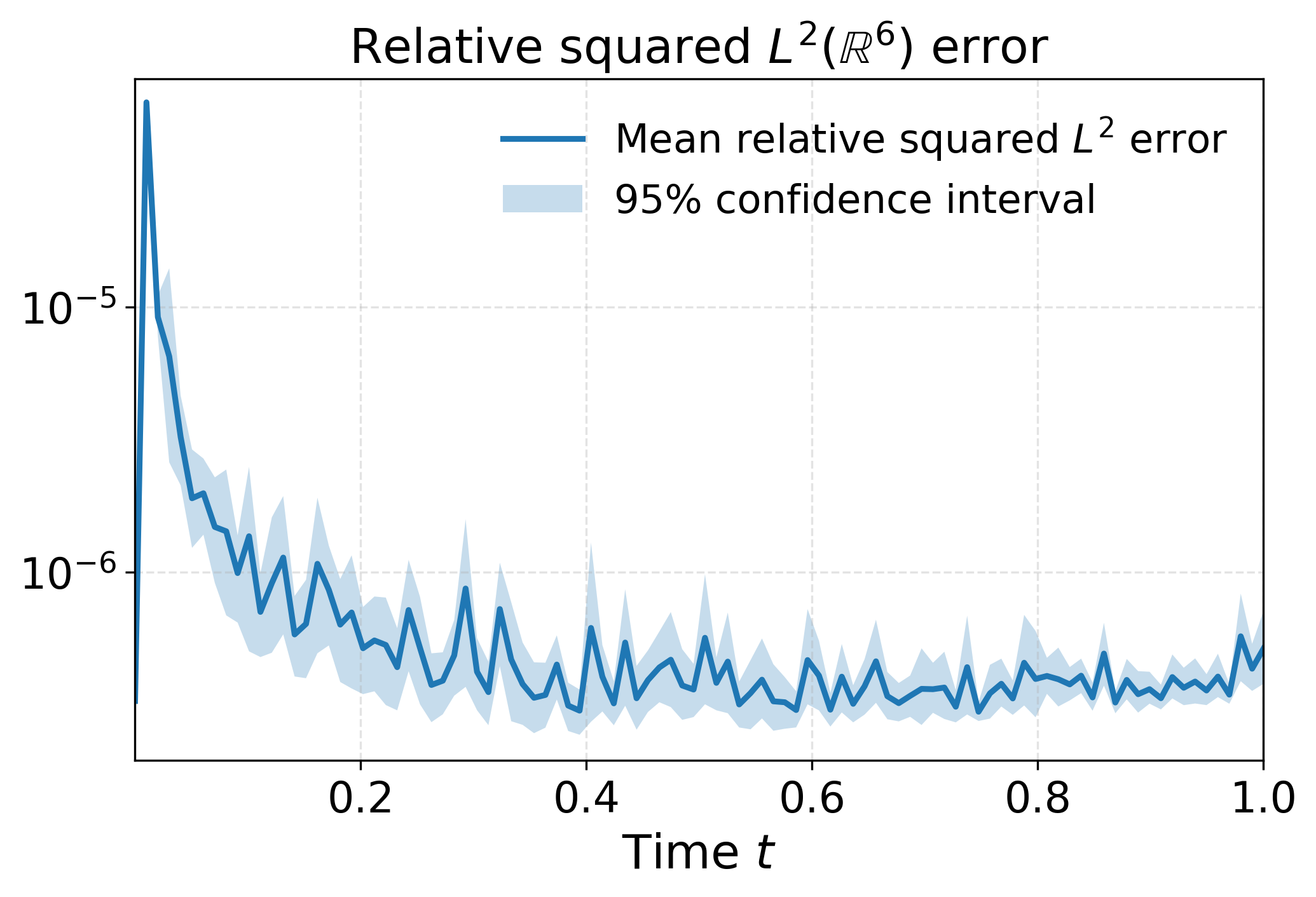}
  \caption{}
\end{subfigure}
\caption{Example of Sect.~\ref{sect:ngnf_proc_shear}. The first two figures show the Normalizing Flow approximation and the absolute error on the $(x_1,x_2)$ section of the TPDF of $\boldsymbol{X}(t)$. The rightmost figure reports, on a logarithmic vertical scale,  the mean relative squared $L^2(\mathbb{R}^6)$ error over time, together with a $95\%$ bootstrap confidence interval for the mean.}
\label{fig:ngnf_proc_nongaussian_anisotropic}
\end{figure}

In Fig.~\ref{fig:ngnf_proc_nongaussian_anisotropic}, left, we show
the two-dimensional section $(x_1,x_2)$ of the Normalizing Flow solution at
$x_3=\cdots=x_6=0$, $\boldsymbol{x_0}=\boldsymbol{0}$, and $t=1$. In Fig.~\ref{fig:ngnf_proc_nongaussian_anisotropic}, right, we compute the relative squared $L^2$ error over time for 1000 samples of $\boldsymbol{x_0}\sim\mathcal{N}_6(\boldsymbol{0},\mathbb{I}_6)$ and display the mean along with the $95\%$ confidence interval. 
Due to the 6-dimensional Dirac delta distribution, the relative squared $L^2$ error is highest immediately after the initial time.
However, its magnitude stays below $10^{-4}$ and decreases 
rapidly over time, 
suggesting that NGNF holds promising potential for addressing high-dimensional problems.

\ethics{Competing Interests}{ 
This study was funded by EPFL and the ETH-Domain Joint Initiative through the UrbanTwin project (\url{https://urbantwin.ch/}).}


\end{document}